\documentclass[cameraready]{Interspeech}
\usepackage{multirow}
\usepackage{graphicx}
\usepackage{subcaption}
\usepackage[table]{xcolor}
\definecolor{mintbg}{rgb}{.63,.79,.95}
\colorlet{lightmintbg}{mintbg!30}
\definecolor{mintbg1}{rgb}{1.0,0.85,0.88}
\colorlet{lightmintbg1}{mintbg1!30}
\usepackage{adjustbox}

\title{Overcoming Decoder Inconsistencies in Whisper for Dravidian and Low-Resource Languages}

\author[equalcontribution]{Chowdam}{Venkata Kumar}
\author[equalcontribution, correspondingauthor]{Kumud}{Tripathi}
\author[]{Pankaj}{Wasnik}

\address{
     Media Analysis Group, Sony Research India 
}

\email{{kumud.tripathi@sony.com, chowdam.kumar@sony.com, pankaj.wasnik}@sony.com}


\usepackage{comment}

\begin{document}

\maketitle

\begin{abstract}
    Multilingual ASR models such as Whisper perform well on high-resource languages but exhibit substantially higher Word Error Rates (WER) for Dravidian languages compared to Indo-Aryan ones. Through linguistic and dataset analysis, we show that Dravidian languages have longer words, higher vocabulary diversity, and lower repetition, resulting in sparse token distributions and frequent character-level substitution errors. Baseline fine-tuning further reveals decoder imbalance between self-attention (linguistic context) and cross-attention (acoustic cues). Although synthetic token-repetition experiments indicate potential gains, they are impractical. Motivated by these observations, we introduce two decoder-level enhancements: Weighted-Attention, which adaptively balances attention sources, and Self-Conditioning, which reinjects intermediate predictions to improve token consistency. Experiments demonstrate consistent WER reductions for low-resource and agglutinative languages.
\end{abstract}

\section{Introduction}

Automatic speech recognition (ASR) has advanced rapidly in recent years, primarily driven by transformer-based architectures such as Whisper. While these models perform well on high-resource languages, they continue to show significantly lower performance on Indo-Aryan and Dravidian languages. Notably, Dravidian languages such as Tamil, Telugu, and Kannada exhibit significantly higher Word Error Rates (WER) compared to Indo-Aryan counterparts like Hindi and Bengali, even when trained under identical conditions.

We begin our investigation by analyzing the linguistic and statistical properties of Indo-Aryan and Dravidian language corpus. Our analysis reveals that Dravidian languages are characterized by longer average word lengths, greater vocabulary diversity, and lower word repetition rates, which together contribute to increased lexical sparsity and increased decoding difficulty. These characteristics pose challenges for both the internal language modeling and the acoustic alignment capabilities of current ASR systems. Further analysis reveals that most errors arise from character-level substitutions, often within known words. This pattern confirms that the decoder struggles to maintain token consistency under complex morphological structures.

Motivated by these findings, we introduce two architectural modifications to Whisper’s decoder: (1) a Weighted-Attention mechanism that adaptively fuses self- and cross-attention signals, and (2) a Self-Conditioning module, where intermediate token predictions are re-injected to guide future decoding steps. Together, these modifications help the decoder maintain consistency and focus in morphologically rich, low-resource language settings. Our experiments show a consistent reduction in WER across multiple Dravidian languages, demonstrating that linguistic-aware decoder design is critical for equitable multilingual ASR performance.

\section{Related Work}
Modern multilingual ASR systems are largely built on Transformer-based encoder-decoder architectures that jointly model acoustic, pronunciation, and language information within a unified framework. Prior work has demonstrated that incorporating subword units along with explicit language-symbol conditioning significantly improves recognition for low-resource languages, yielding around 10\% relative WER reduction \cite{zhou2018multilingual}. Similarly, integrating auxiliary CTC objectives into multilingual training has shown strong gains; conditioning on language identity across 102 languages resulted in a 28\% relative CER reduction \cite{chen2023improving}. These studies highlight the importance of structured conditioning and auxiliary supervision in improving multilingual robustness.

Recent research has also focused on refining decoder attention mechanisms to better integrate acoustic and linguistic cues. For instance, a self-and-mixed attention detector combining acoustic self-attention with linguistic mixed attention achieved state-of-the-art performance on Aishell-1 \cite{zhou2020self}. Dual-decoder architectures with cross-decoder attention between ASR and speech translation pathways further demonstrate the benefits of advanced attention designs for contextual modeling \cite{le2020dual}. Collectively, these approaches emphasize that attention modulation within the decoder plays a critical role in improving recognition accuracy.
Despite such progress, morphologically rich and agglutinative languages, particularly Dravidian languages remain underexplored. Comparative studies using strong pretrained models such as W2V2-BERT \cite{chung2021w2v}, XLSR-53 \cite{conneau2020unsupervised}, and Whisper-small \cite{radford2023robust} consistently report higher WERs and slower convergence for Tamil, Telugu, and Malayalam, attributing the gap to complex phonology, agglutination, and high lexical diversity \cite{jain2025comparative,tripathi2025enhancing}. Similar findings have been reported in Malayalam ASR, where n-gram-based subword modeling significantly influences both memory usage and error rates \cite{manohar2023improving}. These results suggest that vocabulary sparsity and long word formations pose unique challenges not sufficiently addressed by standard multilingual training strategies.

Cross-lingual adaptation techniques provide complementary directions. Parameter-efficient reprogramming of pretrained English Conformer models with only 5\% additional parameters achieves competitive WERs across multiple languages \cite{yang2023english}. Context-conditioning using BERT-derived embeddings with multi-head attention reduces slot errors for rare or domain-specific tokens \cite{djeffal2023automatic}, while retrieval-based k-NN corrections improve long-tail named entity recognition with a 7.4\% relative WER reduction \cite{bekal2021remember, zhang2016introduction}. Although effective, these methods primarily target adaptation or rare-token handling rather than decoder-level imbalance arising from morphological sparsity.

In previous works \cite{wagner2024outlier, bondarenko2023quantizable}, gated attention mechanisms have been explored for post-training quantization and outlier reduction, their objective differs from ours. Our Weighted-Attention dynamically balances self- and cross-attention within the Whisper decoder during training and inference, aiming to mitigate linguistic-acoustic imbalance and reduce character-level substitution errors in low-resource agglutinative languages.
Similarly, prior self-conditioning approaches primarily operate in non-autoregressive encoder frameworks to improve decoding efficiency \cite{komatsu2022non, xiao2023survey, lyu2025nar}. In contrast, our Self-Conditioning mechanism functions within the autoregressive Whisper decoder, reinjecting intermediate predictions to enhance token-level consistency during sequential decoding. To the best of our knowledge, decoder-level conditioning explicitly designed to address morphological sparsity in multilingual ASR remains underexplored.

\begin{table}[t]
\caption{Linguistic and statistical properties of various Indian languages.}\label{tab:language_stats}
\resizebox{\linewidth}{!}
{\begin{tabular}{|l|c|c|c|c|c|c|}
\hline
\multirow{2}{*}{\textbf{Languages}} & \textbf{UC} & \textbf{Duration} & \textbf{Total Words} & \textbf{Avg. WL} & \textbf{Avg. } &\multirow{2}{*}{\textbf{TTR}} \\
 & \textbf{(in K)} & \textbf{(in Hrs)} & \textbf{ (in K)} & \textbf{(in chars)} & \textbf{WR} &  \\
\hline
\rowcolor{lightmintbg}\multicolumn{7}{|c|}{Indo-Aryan Languages}\\
\hline
Hindi      & 93.02  & 137.18 & 1095.22 & 3.98 & 18.93 & 0.05 \\
Gujarati   & 71.06  & 116.22 & 820.70  & 4.81 & 9.13  & 0.11 \\
Marathi    & 97.18  & 172.08 & 1111.67 & 5.83 & 7.88  & 0.13 \\
Bengali    & 57.91  & 102.52 & 628.76  & 5.65 & 10.22 & 0.10 \\ \hline
\rowcolor{lightmintbg}\multicolumn{7}{|c|}{Dravidian Languages}\\
\hline
Tamil      & 95.77  & 171.92 & 1023.09 & 8.07 & 5.50  & 0.18 \\
Telugu     & 70.69  & 141.77 & 767.49  & 6.87 & 5.72  & 0.17 \\
Kannada    & 70.16  & 152.69 & 779.19  & 7.19 & 4.96  & 0.20 \\
Malayalam  & 45.12  & 134.14 & 489.69  & 9.14 & 3.25  & 0.31 \\
\hline
\end{tabular}}
\end{table}

\section{Preliminary Study}
\label{sec:linguistic-analysis}

\subsection{\textbf{Linguistic Variation Analysis}}We perform a corpus-level linguistic analysis using the Kathbath dataset \cite{javed2023indicsuperb}, which includes languages from both Indo-Aryan (Hindi, Gujarati, Marathi, Bengali) and Dravidian (Tamil, Telugu, Kannada, Malayalam) families. The analysis focuses on type-to-token ratio (TTR), average word repetition (WR), and word length (WL), along with dataset duration and utterance counts (UC), to uncover factors influencing multilingual ASR performance.
As shown in Table \ref{tab:language_stats}, the dataset suffers from significant imbalances across languages. For instance, Malayalam, despite having a duration similar to Hindi, has less than half its utterances. This disproportion leads to underrepresentation during training, where languages with fewer utterances are sampled less frequently, reducing their recognition accuracy. Sampling-based strategies \cite{junsomboon2017combining} attempt to mitigate this issue but risk overfitting low-resource languages before high-resource ones are fully learned, underscoring the structural imbalance in multilingual datasets.

Linguistic complexity further compounds the problem. Dravidian languages exhibit TTR values nearly twice those of Indo-Aryan languages, reflecting their agglutinative morphology and greater lexical diversity. Within families, variation persists, Hindi is relatively less complex, whereas Malayalam demonstrates the highest complexity, with long words that extend sequence length and complicate decoding. Together, dataset imbalance and linguistic diversity create dual challenges for end-to-end systems like Whisper, which must simultaneously handle highly varied phonological and morphological structures across languages.

\subsection{Character-level Error Analysis}
To better understand the sources of recognition errors, we conduct a preliminary analysis focusing on substitution patterns within correctly recognized lexical items. In Table \ref{tab:char_error_analysis}, results indicate that a substantial proportion of overall WER originates from character-level substitutions occurring inside known words rather than from insertions or deletions.
This trend suggests that the model often identifies the correct word boundary and general lexical structure but fails to consistently predict the exact character sequence. Such behavior is particularly prominent in morphologically rich languages, where long word forms and dense affixation increase token complexity. The decoder appears to generate partially correct tokens with minor character mismatches, reflecting instability in internal language modeling rather than purely acoustic misalignment.

\begin{table}[t]
\centering
\caption{Known-Unknown word error contribution towards total WER across languages.}
\label{tab:char_error_analysis}
\resizebox{\linewidth}{!}
{\begin{tabular}{|l|c|cc|cc|}
\hline
\multirow{2}{*}{\textbf{Language}} & \textbf{Total} & \multicolumn{2}{c|}{\textbf{Known Words}}  & \multicolumn{2}{c|}{\textbf{Unknown Words}} \\ \cline{3-6}
 & \textbf{WER} & \textbf{WER} & \textbf{ Substitution} & \textbf{WER} & \textbf{Substitution} \\ 
\hline
\rowcolor{lightmintbg}\multicolumn{6}{|c|}{Indo-Aryan Languages}\\
\hline
Hindi      & 10.74 & 8.67  & 7.06  & 2.07  & 1.95 \\
Gujarati   & 16.94 & 13.39 & 11.31 & 3.55  & 3.28 \\
Marathi    & 15.91 & 12.02 & 9.63  & 3.89  & 3.58 \\
Bengali    & 12.55 & 10.31 & 8.37  & 2.24  & 2.07 \\
\hline
\rowcolor{lightmintbg}\multicolumn{6}{|c|}{Dravidian Languages}\\
\hline
Tamil      & 23.85 & 16.68 & 12.58 & 7.17  & 6.38 \\
Telugu     & 23.31 & 16.80 & 12.81 & 6.51  & 5.83 \\
Kannada    & 19.06 & 13.56 & 10.38 & 5.50  & 4.96 \\
Malayalam  & 35.98 & 24.08 & 19.56 & 11.91 & 10.61 \\
\hline
\end{tabular}}
\end{table}

\begin{table}[t]
\caption{WER comparison before and after morpheme splitting using IndicNLP\cite{indicnlp}, 
and vocabulary reduction statistics. MS: Morphological Splitting, UW: Unique Words, UMU: Unique Morphological Unit}
\label{tab:wer_morpheme_comparison}
\resizebox{\linewidth}{!}
{\begin{tabular}{|l|ccc|c|c|c|}
\hline
\multirow{2}{*}{\textbf{Languages}} & \multicolumn{3}{c|}{\textbf{Whisper-Medium}} & \textbf{UW} & \textbf{UMU} & {\textbf{Reduction}} \\ \cline{2-4}
 & \textbf{FT} & \textbf{FT with MS} & \textbf{Imp.} & \textbf{ (in K)} & \textbf{(in K)} & \textbf{(in K)} \\
\hline
\rowcolor{lightmintbg}\multicolumn{7}{|c|}{Indo-Aryan Languages}\\
\hline
Hindi      & 10.74 & 11.48 & -0.74 & 57.85  & 22.82 & 35.02 \\
Gujarati   & 16.94 & 15.89 &  1.05 & 89.81  & 25.11 & 64.69 \\
Marathi    & 15.91 & 15.45 &  0.46 & 140.95 & 26.83 & 114.12 \\
Bengali    & 12.55 & 11.03 &  1.52 & 60.28  & 15.67 & 44.61 \\\hline
\rowcolor{lightmintbg}\multicolumn{7}{|c|}{Dravidian Languages}\\
\hline
Tamil      & 23.85 & 19.82 &  4.03 & 185.93 & 43.93 & 141.99 \\
Telugu     & 23.31 & 18.73 &  4.58 & 134.10 & 31.34 & 102.76 \\
Kannada    & 19.06 & 17.15 &  1.91 & 156.80 & 47.19 & 109.61 \\
Malayalam  & 35.98 & 27.89 &  8.09 & 150.31 & 21.41 & 128.91 \\
\hline
\rowcolor{lightmintbg1}\textbf{Average} & \textbf{19.79} & \textbf{17.18} & \textbf{2.61} & -- & -- & -- \\
\hline
\end{tabular}}
\end{table}

\subsection{\textbf{Morphological Analysis}}
Our study investigates the impact of morphological complexity on multilingual ASR performance, particularly for low-resource agglutinative languages. Dravidian languages exhibit rich inflectional structures and long word formations, resulting in large vocabularies with low token repetition and pronounced data sparsity. This sparsity limits the model’s ability to learn stable lexical representations. As shown in Table \ref{tab:wer_morpheme_comparison}, morpheme-level segmentation mitigates this issue by decomposing words into smaller, reusable units, thereby reducing vocabulary size and improving generalization. Substantial WER reductions are observed for Malayalam (8.09\%), Telugu (4.58\%), and Tamil (4.03\%), with an average improvement of 2.61\%. Although Hindi shows slight degradation, the overall trend confirms that morphological sparsity significantly contributes to recognition errors in agglutinative languages. Importantly, morpheme splitting is employed purely as an analytical tool rather than a standalone contribution. Since these improvements rely on post-processing, they motivate our core objective: incorporating morphological awareness directly into Whisper’s decoder through Weighted-Attention and Self-Conditioning for more robust multilingual ASR.

\begin{figure}[ht]
    \centering
    \begin{subfigure}[t]{0.2\textwidth}
        \centering
        \includegraphics[width=\linewidth]{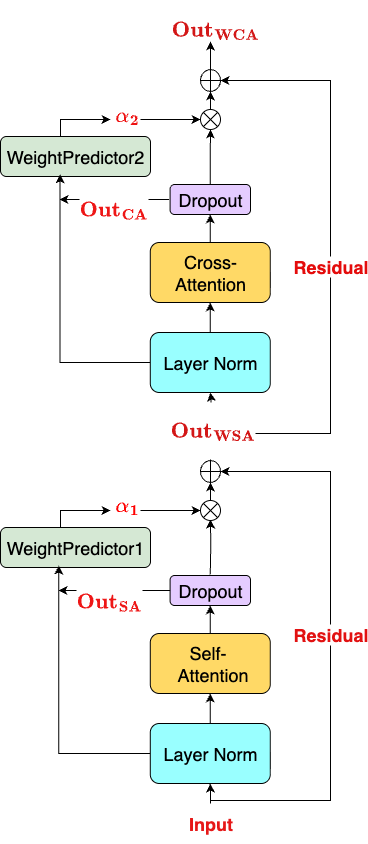}
        \caption{Weighted-Attention: Adaptive gating balances self- and cross-attention within decoder layer.}
        \label{fig:fig1}
    \end{subfigure}
    \hfill
    \begin{subfigure}[t]{0.25\textwidth}
        \centering
        \includegraphics[width=\linewidth]{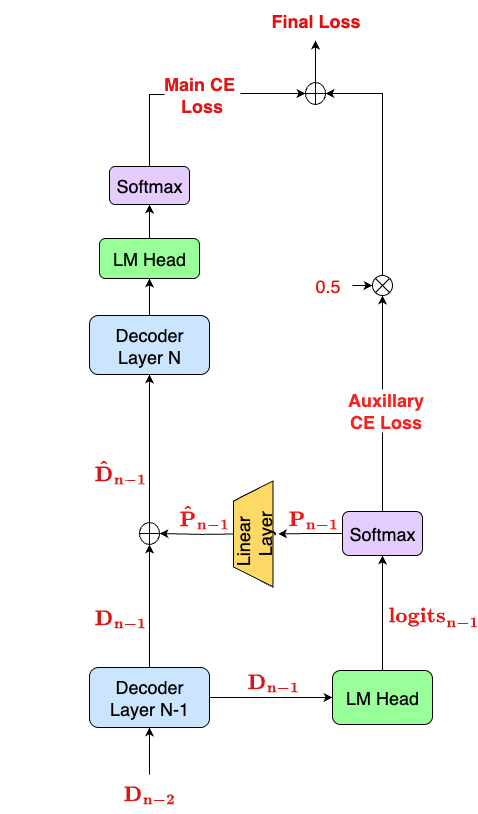}
        \caption{Self-Conditioning: Intermediate decoder predictions reinjected with auxiliary loss for consistency.}
        \label{fig:fig2}
    \end{subfigure}
    \caption{Block diagrams of proposed approaches.}
    \label{fig:block}
\end{figure}

\begin{table*}[t]
\centering
\caption{WER comparison across languages for different methods: Base Whisper, Weighted-Attention, Self-Conditioning, and their combination. Morphological splitting (MS) is applied for improved recognition.}
\label{tab:wer_comparison}
\begin{adjustbox}{width=0.9\textwidth}
\begin{tabular}{|l|l|c|c|c|c|c|c|c|c|c|}
\hline
\multicolumn{2}{|c|}{\textbf{Model with FT Approach}} & \textbf{Hi} & \textbf{Gu} & \textbf{Mr} & \textbf{Bn} & \textbf{Ta} & \textbf{Te} & \textbf{Kn} & \textbf{Ma} & \textbf{Average} \\
\hline

{\multirow{3}{*}{\textbf{W-M (Baseline)}} }
& FT & 10.74 & 16.94 & 15.91 & 12.55 & 23.85 & 23.31 & 19.06 & 35.98 & 19.79 \\
 & FT with MS & 11.48 & 15.89 & 15.45 & 11.03 & 19.82 & 18.73 & 17.15 & 27.89 & 17.18 \\
& \cellcolor{lightmintbg}Improvement & \cellcolor{lightmintbg}\textbf{-0.74} & \cellcolor{lightmintbg}\textbf{1.05} & \cellcolor{lightmintbg}\textbf{0.46} & \cellcolor{lightmintbg}\textbf{1.52} & \cellcolor{lightmintbg}\textbf{4.03} & \cellcolor{lightmintbg}\textbf{4.58} & \cellcolor{lightmintbg}\textbf{1.91} & \cellcolor{lightmintbg}\textbf{8.09} & \cellcolor{lightmintbg1}\textbf{2.61} \\
\hline
{\multirow{3}{*}{\textbf{+\,Weighted-Attention}} }
& FT & 10.26 & 16.05 & 14.38 & 11.24 & 22.80 & 21.76 & 18.44 & 34.05 & 18.62 \\
 & FT with MS  & 10.67 & 14.61 & 13.65 & 9.31 & 18.52 & 16.69 & 16.07 & 25.54 & 15.63 \\
& \cellcolor{lightmintbg}Improvement & \cellcolor{lightmintbg}\textbf{0.81} & \cellcolor{lightmintbg}\textbf{1.28} & \cellcolor{lightmintbg}\textbf{1.80} & \cellcolor{lightmintbg}\textbf{1.72} & \cellcolor{lightmintbg}\textbf{1.30} & \cellcolor{lightmintbg}\textbf{2.04} & \cellcolor{lightmintbg}\textbf{1.08} & \cellcolor{lightmintbg}\textbf{2.35} & \cellcolor{lightmintbg1}\textbf{1.54} \\
\hline
{\multirow{3}{*}{\textbf{+\,Self-Conditioning}} }
& FT & 10.40 & 16.24 & 14.70 & 11.20 & 22.99 & 21.68 & 17.83 & 33.55 & 18.57 \\
& FT with MS  & 10.82 & 14.66 & 13.97 & 9.22 & 18.72 & 16.90 & 15.68 & 25.07 & 15.63 \\
& \cellcolor{lightmintbg}Improvement & \cellcolor{lightmintbg}\textbf{0.66} & \cellcolor{lightmintbg}\textbf{1.23} & \cellcolor{lightmintbg}\textbf{1.48} & \cellcolor{lightmintbg}\textbf{1.81} & \cellcolor{lightmintbg}\textbf{1.10} & \cellcolor{lightmintbg}\textbf{1.83} & \cellcolor{lightmintbg}\textbf{1.47} & \cellcolor{lightmintbg}\textbf{2.82} & \cellcolor{lightmintbg1}\textbf{1.55} \\
\hline
{\multirow{3}{*}{\textbf{+\,Combined}} }
& FT & 10.35 & 16.00 & 14.42 & 11.17 & 22.83 & 21.61 & 18.02 & 33.41 & \cellcolor{lightmintbg1}\textbf{18.48} \\
& FT with MS  & 10.48 & 14.62 & 13.61 & 9.33 & 18.77 & 16.70 & 15.80 & 24.89 & \cellcolor{lightmintbg1}\textbf{15.53} \\
& \cellcolor{lightmintbg}Improvement & \cellcolor{lightmintbg}\textbf{1.00} & \cellcolor{lightmintbg}\textbf{1.27} & \cellcolor{lightmintbg}\textbf{1.84} & \cellcolor{lightmintbg}\textbf{1.70} & \cellcolor{lightmintbg}\textbf{1.05} &\cellcolor{lightmintbg}\textbf{ 2.03} & \cellcolor{lightmintbg}\textbf{1.35} & \cellcolor{lightmintbg}\textbf{3.00} & \cellcolor{lightmintbg1}\textbf{1.65 }\\
\hline
\end{tabular}
\end{adjustbox}
\end{table*}

\section{Proposed Methodology}

In this section, we present details of the proposed Weighted-Attention mechanism to balance linguistic and acoustic cues adaptively, and a Self-Conditioning module to reinforce token consistency during decoding.

\subsection{Weighted-Attention}
The main challenge lies in the decoder’s difficulty in balancing acoustic cues from cross-attention with contextual cues from self-attention, often resulting in phoneme confusions and recognition errors in Dravidian languages. To address this, we introduce two lightweight gating modules in each decoder layer: WeightPredictor1 for self-attention and WeightPredictor2 for cross-attention. Each module consists of a two-layer feed-forward network with sigmoid activation, producing scores in the range $[0,1]$. These scores dynamically regulate attention contributions, with weighted outputs integrated via residual connections ($Res$), as shown in Figure \ref{fig:block}(a).
\begin{align}
\alpha_1 = \text{WeightPredictor1}( concat({Res},Out_{SA} )) \nonumber \\
Out_{WSA} = Res + \alpha_1 \cdot Out_{SA} \\
\alpha_2 = \text{WeightPredictor2}(concat({Out_{WSA}},Out_{CA} )) \nonumber \\
    Out_{WCA} = Out_{WSA} + \alpha_2 \cdot Out_{CA}
\end{align}

\noindent Here, $\alpha_1$ and $\alpha_2$ describe a gating mechanism where WeightPredictor1 generates $\alpha_1$ to regulate self-attention ($Out_{SA}$ → $Out_{WSA}$), and WeightPredictor2 computes $\alpha_2$ to balance cross-attention ($Out_{CA}$ → $Out_{WCA}$), adaptively refining decoder outputs.

\subsection{Self-Conditioning}
The second challenge involves handling long sequences and extracting meaningful patterns in deeper layers, which is harder in low-resource languages with limited acoustic and contextual cues. To address this, we adopt a self-conditioning mechanism applied at the second-last decoder layer, which we found to be the most effective after experimenting with multiple layers. It re-injects partial predictions ($P$) as guiding signals, refining outputs and emphasizing relevant linguistic details. 

We further conducted an ablation study to examine the individual contributions of self-conditioning and the auxiliary cross-entropy loss. The results show that each component provides moderate improvements independently, while their combination consistently delivers superior performance. Therefore, in subsequent experiments, we report results using the combined self-conditioning framework with auxiliary supervision. When integrated with adaptive attention weighting, this approach significantly enhances accuracy and robustness in multilingual ASR, particularly for low-resource and agglutinative languages.
\begin{align}
    logits_{n-1} &= \text{LMHead}(D_{n-1}) \nonumber \\
    P_{n-1} &= \text{softmax}(logits_{n-1})  \\
    \hat{P}_{n-1} &= \text{linear}(P_{n-1}) \nonumber \\
    \hat{D}_{n-1} &= D_{n-1} + \hat{P}_{n-1}
\end{align}

\noindent These equations illustrate prediction feedback in the decoder. The state $D_{n-1}$ generates logits via the LM head, transformed by softmax into probabilities $P_{n-1}$. These predictions are linearly projected to $\hat{P}_{n-1}$, aligning them with the decoder’s space. Finally, $\hat{P}_{n-1}$ is added back to $D_{n-1}$, producing $\hat{D}_{n-1}$, a richer, context-integrated representation.

\section{Experimental setup}

\subsection{Implementation Details}
Experiments were conducted on four NVIDIA A100 (40GB) GPUs using Whisper-medium. Fine-tuning was performed for 3 epochs with the AdamW optimizer, a batch size of 16, and learning rates of 1e-5 for standard fine-tuning and 5e-5 for newly introduced parameters. All experiments were implemented using the Hugging Face Transformers toolkit, ensuring reproducibility and consistency across settings.

\subsection{Dataset and Evaluation Metrics}
 This study primarily utilizes the Indian multilingual dataset comprising both Indo-Aryan languages: Hindi (Hi), Gujarati (Gu), Marathi (Mr), Bengali (Bn), and Dravidian languages: Tamil (Ta), Telugu (Te), Kannada (Kn), Malayalam (Ma), ensuring balanced representation across language families. The datasets were sourced from the publicly available Kathbath speech corpus. For consistency, we adopted the same training, validation, and testing splits as described in \cite{javed2023indicsuperb}. To evaluate cross-lingual generalization, we further conduct experiments on two non-Indian agglutinative languages: Korean \cite{openslr40} and Swahili \cite{mozilla_commonvoice}, following the respective standard splits provided in \cite{openslr40} and \cite{mozilla_commonvoice}.
 The performance of the model was evaluated using Word Error Rate (WER in \%).

\section{Results and discussion}
\subsection{Analysis on Indian Languages}
The results in Table \ref{tab:wer_comparison} show the relative improvements of different decoder enhancement methods over the baseline Whisper-medium fine-tuning (W-M FT) with Morphological Splitting (MS). On average, Weighted-Attention and Self-Conditioning yield comparable overall gains (1.54\% and 1.55\%), improving recognition accuracy across both Indo-Aryan and Dravidian languages. The gains are modest for relatively easier languages like Hindi (0.81–0.66\%) but more pronounced for morphologically complex Dravidian languages, such as Malayalam (2.35–2.82\%) and Telugu (2.04–1.83\%). The combined method achieves the highest relative improvements for some Dravidian languages, including Malayalam (3.00\%) and Telugu (2.03\%), though it provides only a slight average increase (1.65\%) over individual methods. These findings confirm that decoder-level conditioning is particularly beneficial for high-morphology, high-WER languages, aligning with our focus on Dravidian ASR challenges.
Importantly, both Weighted-Attention and Self-Conditioning are lightweight enhancements, introducing less than 1\% additional parameters through small feed-forward networks. The training overhead is minimal (2–3\%), with no noticeable increase in peak GPU memory usage. Inference latency rises by less than 2\%, as the operations involve simple linear projections and gating mechanisms.
We also conducted experiments on other Whisper variants, including Whisper-small and Whisper-large-v3, and observed consistent performance improvement (1-2\%) with our proposed methods.

\begin{table}[t]
\centering
\caption{WER comparison on non-Indian Agglutinative languages using baseline and proposed approaches.}
\label{tab:models_list}
\begin{adjustbox}{width=0.35\textwidth}
\begin{tabular}{|l|c|c|}
\hline
\textbf{Model} & \textbf{Korean} & \textbf{Swahili}\\ 
\hline
\textbf{W-M FT (Baseline)} & 3.34 & 16.07\\
\textbf{+\,Weighted-Attention} & 2.86 & 15.48\\
\textbf{+\,Self-Conditioning} & 2.77 & 15.12\\
\textbf{+\,Combined}  & \cellcolor{lightmintbg}\textbf{2.51} & \cellcolor{lightmintbg}\textbf{14.58}\\
\hline
\end{tabular}
\end{adjustbox}
\end{table}

\subsection{Ablation on non-Indian Agglutinative Languages}
Table \ref{tab:models_list} reports the performance comparison on two non-Indian agglutinative languages, Korean and Swahili, across different models using FT without MS. These languages were specifically chosen because, similar to Dravidian languages, they possess rich morphological structures and highly agglutinative characteristics, often resulting in long and complex word forms. Such properties make them well-suited for assessing the broader applicability of our methods outside the Indian language context. For both Korean and Swahili, all models were trained separately. The baseline system records a WER of 3.34\% for Korean and 16.07\% for Swahili. Introducing the weighted-attention mechanism leads to clear gains, lowering WER to 2.86\% and 15.48\%, respectively. The self-conditioning approach further improves performance, achieving 2.77\% for Korean and 15.12\% for Swahili. The most significant improvements arise when both strategies are combined, with WER reduced to 2.51\% for Korean and 14.58\% for Swahili. These consistent improvements confirm the robustness of the proposed techniques and demonstrate their effectiveness in generalizing to morphologically complex, low-resource agglutinative languages beyond the Indian setting.

\section{Conclusion}
This work investigates the persistent performance gap between Dravidian and Indo-Aryan languages in multilingual ASR systems, focusing on Whisper. Through corpus analysis and decoding behavior, we identify high character-level substitution errors in Dravidian languages, driven by complex morphology, longer words, and low repetition. To address this, we propose two decoder-level enhancements: a Weighted-Attention mechanism to balance self- and cross-attention, and a Self-Conditioning module that feeds back intermediate predictions to guide decoding. Experimental results show consistent WER improvements, particularly in known substitution cases, validating our hypothesis. These findings highlight the need for language-aware decoder conditioning and suggest effective architectural strategies to improve ASR for morphologically rich, low-resource languages like Tamil and Telugu.

\section{Generative AI Use Disclosure}
Generative AI tools have been used mainly for grammar correction, paraphrasing, and overall language editing purposes in the manuscript. All scientific contributions, technical implementations, and interpretations were developed and validated by the authors.
In accordance with policy guidelines, no generative AI system is listed as a co-author, and all authors take full responsibility and accountability for the content and integrity of this paper.

\bibliographystyle{IEEEtran}
\bibliography{mybib}

\end{document}